\documentclass{article}

\usepackage{arxiv}

\usepackage[utf8]{inputenc} 
\usepackage[T1]{fontenc}    
\usepackage{hyperref}       
\usepackage{url}            
\usepackage{booktabs}       
\usepackage{amsfonts}       
\usepackage{nicefrac}       
\usepackage{microtype}      
\usepackage{lipsum}
\usepackage{graphicx}
\usepackage{subfig}
\graphicspath{ {./images/} }

\title{Autoencoding Generative Adversarial Networks}

\author{
  Conor Lazarou \\
  Flatland Data Solutions\\
  Saskatoon, Canada \\
  \texttt{conor@flatland.ai} \\
}

\begin{document}

\maketitle

\begin{figure}[h] 
    \centering
    \includegraphics[width=16cm]{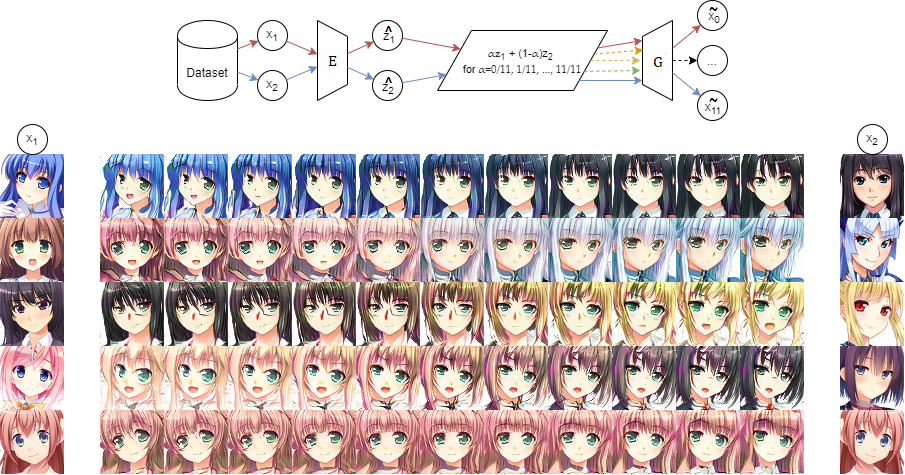}
    \caption{Direct interpolation between real samples. (Top) The linear interpolation process. Two samples $x_1$ and $x_2$ are selected and encoded by $E$ into latent vectors $\hat{z}_1$ and $\hat{z}_2$. Intermediate vectors are linearly interpolated between those vectors, and the interpolated vectors are used by the generator $G$ to produce interpolated samples. (Bottom) The results of the above process for five pairs of real samples. The bottom pair of samples is a single image flipped horizontally.}
    \label{fig:interpolation}
\end{figure}

\begin{abstract}
In the years since Goodfellow \textit{et al.} introduced Generative Adversarial Networks (GANs) \cite{goodfellow2014generative}, there has been an explosion in the breadth and quality of generative model applications. Despite this work, GANs still have a long way to go before they see mainstream adoption, owing largely to their infamous training instability. Here I propose the Autoencoding Generative Adversarial Network (AEGAN), a four-network model which learns a bijective mapping between a specified latent space and a given sample space by applying an adversarial loss and a reconstruction loss to both the generated images and the generated latent vectors. The AEGAN technique offers several improvements to typical GAN training, including training stabilization, mode-collapse prevention, and permitting the direct interpolation between real samples. The effectiveness of the technique is illustrated using an anime face dataset.
\end{abstract}

\section{Introduction}
For as long as humans have drawn breath we have been captivated by the arts. Cave paintings and jewelry are among the oldest artifacts of our species. Equally ancient and lasting is our chronic laziness and derision of work. In recent years, these two passions have been combined in the field of generative modelling, which allows for the effortless generation of works in a multitude of domains. At long last, we can create art without the drudgery of getting out of bed (assuming one owns a laptop). However, this field is still in its infancy. GANs \cite{goodfellow2014generative}, which have demonstrated photo-realistic results in image generation \cite{zhu2017unpaired, karras2018stylebased, brock2018large}, are notoriously difficult to train \cite{arjovsky2017principled, roth2017}. Because they learn arbitrary, surjective mappings from a latent space (the random noise used in sample generation) to a sample space (the dataset from which we are trying to generate new samples), they are prone to mode collapse \cite{thanhtung2018catastrophic} and do not lend well to analysis of structures within that latent space.

In this paper, I outline the Autoencoding Generative Adversarial Network (AEGAN), which can be seen as a combination of two GANs and two autoencoders (AEs), similar to the CycleGAN formulation \cite{zhu2017unpaired}. The AEGAN technique is a scaffolding for a GAN which learns a bijective mapping between the sample space and the specified latent space. The benefits of this technique are the primary contributions of this paper, and they are threefold:
\begin{itemize}
  \item The AEGAN technique stabilizes GAN training by directing the generator towards the low-dimensional manifold in high-dimensional pixel space.
  \item The AEGAN technique mitigates mode collapse by requiring that the generator be able to reproduce the training dataset.
  \item The AEGAN technique allows for direct interpolation between real samples.
\end{itemize}

\section{Background}
\subsection{Autoencoders}
Autoencoders are a class of self-supervised neural networks with an hourglass shape which learn an efficient, domain-specific encoding of a given dataset drawn from some sample space $X$ \cite{kramer_1991}. Autoencoders consist of two networks: an encoder, $E$, which learns a mapping $E: X \rightarrow Z$, where $Z$ is some lower-dimensional latent space; and a decoder, $G$, which learns a mapping $G: Z \rightarrow X$. The autoencoder is tasked with encoding then reconstructing each sample from the dataset such that $G(E(x))=\widetilde{x} \approx x$, and is penalized with a reconstruction loss. This reconstruction loss is typically some function of the pixel-wise difference between the input $x$ and output $\widetilde{x}$. After fitting, the encoder can be separated, frozen, and used for other purposes, such as in feature extraction for downstream applications \cite{Ryu2020}. Unfortunately, the decoder can't be used as a generative model \cite{foster_2019}. This is because the encodings $E(X)$ are not constrained to a smooth, well-behaved distribution, and as such most points in $Z$ do not correspond to a realistic sample in $X$ \cite{foster_2019} and because pixel-wise reconstruction loss is known to result in blurry samples \cite{larsen2015autoencoding}. The first problem has been addressed by others, namely through Variational Autoencoders (VAEs) \cite{kingma2013autoencoding} and Adversarial Autoencoders (AAEs) \cite{makhzani2015adversarial}, and the latter has been addressed by the VAE/GAN technique \cite{larsen2015autoencoding}. The AEGAN technique proposed in this paper can be considered a combination of the AAE and the VAE/GAN techniques, and as such addresses both of these problems.

\subsection{Generative Adversarial Networks}
GANs are a class of generative, unsupervised neural networks which approximate the data generating distribution $p_{data}$ used to produce a given dataset $X$ \cite{goodfellow2014generative}. They consist of two networks, a generator $G$ and discriminator $D_x$. The generator learns a mapping $G: Z \rightarrow X$, from some latent space $Z$ to the desired sample space $X$, where latent samples are drawn from Z with the distribution $p(z)$ (typically a spherical, 100-dimensional normal distribution). The discriminator is tasked with discerning which samples are drawn from the true distribution (ie, the dataset) and which were generated by $G$. The generator and discriminator take turns updating in a minimax game, with the generator attempting to minimize Equation 1 while the discriminator attempts to maximize it.

\begin{equation}
\min_{G}\max_{D}[\E_{x \sim p_{data}}[\log D_{x}(x)] + \E_{z \sim p(z)}[\log (1 - D_{x}(G(z))]]
\end{equation}

This is equivalent to tasking the generator with minimizing the Jensen-Shannon divergence between $X$ and $G(Z)$ \cite{goodfellow2014generative}. However, the sample space is typically a small manifold in a much larger pixel space; this results in disjoint support between $G(Z)$ and $X$, which results in a perfect (ie, useless) discriminator which provides unhelpful feedback to the generator, leading to lack of convergence and mode collapse \cite{roth2017}. The AEGAN technique addresses this issue.

\subsection{CycleGAN}
CycleGAN is a technique for unpaired image-to-image translation whereby a bijective function is learned, mapping two image sample spaces, $X$ and $Y$, to each other \cite{zhu2017unpaired}. This is accomplished through the use of two GANs, making use of both the adversarial loss (equation 1) and a cycle-consistency loss (ie, a reconstruction loss). The proposed AEGAN technique is a generalization of CycleGAN, wherein one of the sample spaces is the latent space $Z$ and the other is the desired sample space (not necessarily images, although incidentally an image dataset was used for this paper).

\begin{figure}[t] 
    \centering
    \includegraphics[width=9cm]{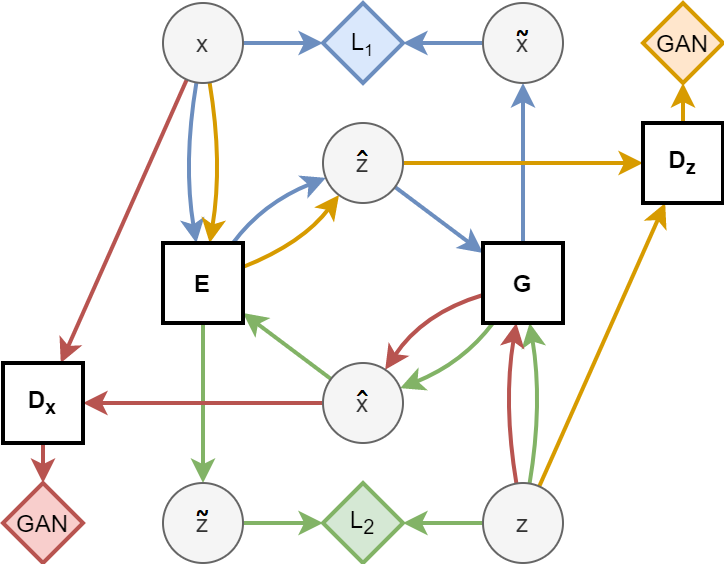}
    \caption{The high-level AEGAN architecture. The four networks are represented as boxes, samples/vectors are represented as circles, and loss functions as diamonds. The colours demonstrate the constituent models: (red) the image-generating GAN, (blue) the image autoencoder, (yellow) the latent vector-generating GAN, (green) the latent vector autoencoder.}
    \label{fig:network}
\end{figure}

\section{Autoencoding Generative Adversarial Networks}
AEGAN is a technique for learning a bijective mapping between some sample space $X$ and a latent space $Z$. An encoder network $E$ is trained to learn a function $E : X \rightarrow Z$, mapping each real sample to a point in the latent space. The generator network $G$ is trained to learn a function $G : Z \rightarrow X$ mapping each point in the latent space to a sample in the sample space. These networks are trained in tandem with two discriminator networks, $D_x$ and $D_z$, which learn to discriminate between real and generated samples and latent points, respectively. The high-level network is illustrated in Figure \ref{fig:network}.

\subsection{Adversarial Loss}
The adversarial loss has four components: we consider generated samples $G(z) = \hat{x}$, encoded samples $E(x) = \hat{z}$, autoencoded samples $G(E(x)) = \widetilde{x}$, and autoencoded latent vectors $E(G(z)) = \widetilde{z}$, where $x$ are real samples drawn from the dataset and $z$ are latent vectors drawn randomly from some distribution. Thus, the four components to the adversarial loss are:

\begin{equation}
\mathcal{L}_{GAN_{\hat{x}}}(G, D_x) = \E_{x \sim p_{data}}[\log D_{x}(x)] + \E_{z \sim p(z)}[\log (1 - D_{x}(G(z))]
\end{equation}
\begin{equation}
\mathcal{L}_{GAN_{\tilde{x}}}(G, E, D_x) = \E_{x \sim p_{data}}[\log D_{x}(x)] + \E_{x \sim p_{data}}[\log (1 - D_{x}(G(E(x)))]
\end{equation}
\begin{equation}
\mathcal{L}_{GAN_{\hat{z}}}(E, D_z) = \E_{x \sim p(z)}[\log D_{z}(z)] + \E_{z \sim p_{data}}[\log (1 - D_{z}(E(x))]
\end{equation}
\begin{equation}
\mathcal{L}_{GAN_{\tilde{z}}}(G, E, D_z) = \E_{x \sim p(z)}[\log D_{z}(z)] + \E_{x \sim p(z)}[\log (1 - D_{z}(E(G(z)))]
\end{equation}

Which, summed, form the adversarial loss ${L}_{GAN}(G, E, D_x, D_z)$.

\subsection{Reconstruction Loss}
Because the goal is to learn a bijective mapping, it is important that $E$ undoes $G$ and vice-versa. Thus, a reconstruction loss is used:
\begin{equation}
\mathcal
{L}_{r}(G, E) = \lambda_{rx}\E_{x \sim p_{data}}[\left| \left| G(E(x)) - x \right| \right|_1] + \lambda_{rz}\E_{z \sim p(z)}[\left| \left| E(G(z)) - z \right| \right|_2] 
\end{equation}
where $\lambda_{rx}$ and $\lambda_{rz}$ are hyperparamters for controlling the weight of the sample reconstruction and latent vector reconstruction, respectively. The $L_1$ loss is used for sample reconstruction because it is known to result in less bluriness than $L_2$ in image reconstruction. 

\subsection{AEGAN Loss}
The full AEGAN objective is thus:
\begin{equation}
\mathcal
{L}_{AEGAN}(G, E, D_x, D_z) = \; {L}_{GAN}(G, E, D_x, D_z) + {L}_{r}(G, E) 
\end{equation}
$G$ and $E$ attempt to minimize ${L}_{AEGAN}$ while $D_x$ and $D_z$ attempt to maximize it.

\begin{figure}[t]%
    \centering
    \subfloat[GAN]{{\includegraphics[width=7.5cm]{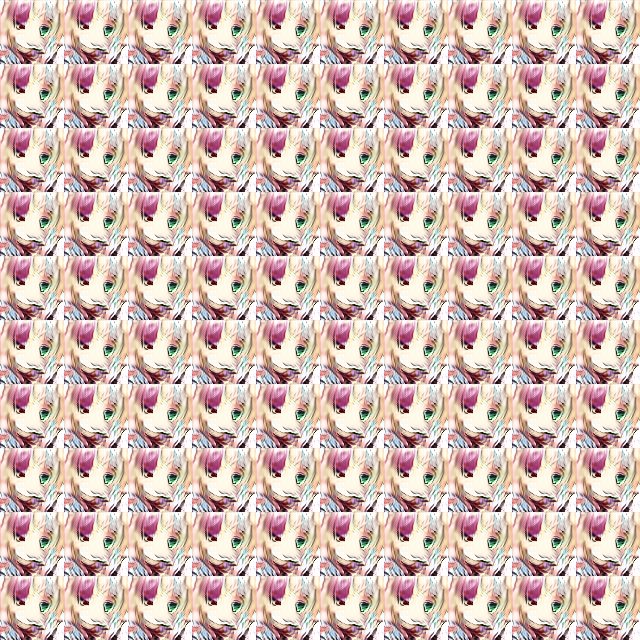} }}%
    \qquad
    \subfloat[AEGAN]{{\includegraphics[width=7.5cm]{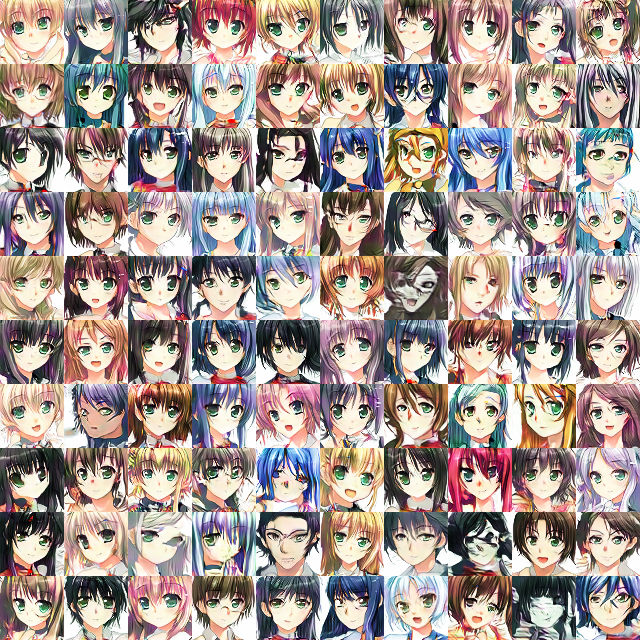} }}%
    \caption{(A) samples generated by a GAN using 100 random vectors and (B) samples generated by an AEGAN using the same random vectors, both networks having been trained for 300000 steps on the anime faces dataset. While GAN training resulted in total mode collapse and low-quality samples, the AEGAN with the exact same generator architecture produced realistic samples with substantial diversity.}%
    \label{fig:generated}%
\end{figure}

\section{Results}

A set of 43104 anime face images (21552 unique faces, augmented by mirroring horizontally), each with a resolution of 64x64 pixels, was used as the sample dataset $X$ \cite{rakshit_2019}, while random vectors $Z$ were drawn from the $N_{32}(0, I)$ distribution. An AEGAN was trained using this data. For comparison, a GAN (comprising of networks $G$ and $D_x$ in Figure \ref{fig:network}) and an AAE (comprising of networks $G$, $E$, and $D_z$ in Figure \ref{fig:network}), both subsets of the AEGAN, were built and trained with the exact same applicable architectures and hyperparameters. Each was trained for 300k training steps (11134 epochs with a minibatch size of 16). Code and hyperparameters used for building and training the models are available at https://github.com/ConorLazarou/AEGAN-keras.

\subsection{Generation}
Random samples generated by the GAN and by the AEGAN are illustrated in Figure \ref{fig:generated}. Not only did the AEGAN's generator avoid mode collapse, unlike the GAN, but it also resulted in substantially higher-quality images, many of which could easily be mistaken for hand-drawn faces. There is substantial diversity in the generated samples, with the exception of eye colour which is overwhelmingly green. Visualization of generated samples during the training process revealed that the AEGAN cycled through various eye colours during training, a common GAN oscillation behaviour. To verify that the stark differences in results were not a fluke, the AEGAN and the GAN were both trained from random initialization for 300k training steps five additional times; in each case, the GAN underwent total mode collapse while the AEGAN did not.
 
\subsection{Reconstruction}
\begin{figure}[t]%
    \centering
    \subfloat[AAE]{{\includegraphics[width=7.5cm]{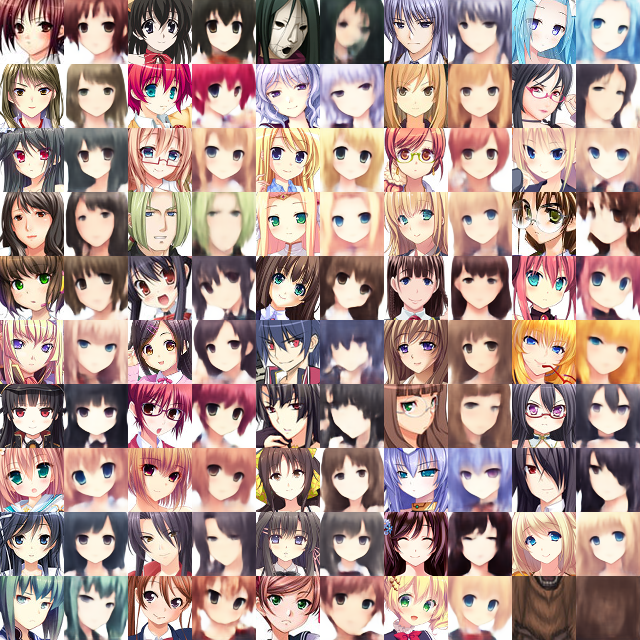} }}%
    \qquad
    \subfloat[AEGAN]{{\includegraphics[width=7.5cm]{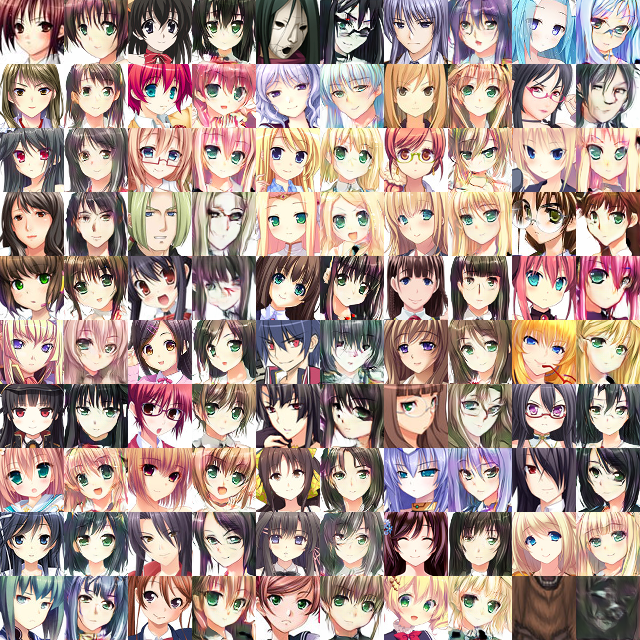} }}%
    \caption{Real samples and their reconstructions from an AAE (a) and an AEGAN (b). Samples are in pairs horizontally, where the left in each pair is the original sample and the right is the reconstruction. Note how the AAE produces blurry samples, while the AEGAN does not. The bottom right appears to be a werewolf's neck; understandably, the reconstruction is terrifying.}%
    \label{fig:reconstructed}%
\end{figure}

Samples from the original dataset and their reconstructions are illustrated in Figure \ref{fig:reconstructed}. The AEGAN was able to effectively reconstruct many of the original images without the telltale bluriness that autoencoders are known for, despite the pixel-wise reconstruction loss. Hair colour, face shape, eye shape, and pose are reproduced reasonably well in most cases. Some shortcomings are obvious, however: there is little variety in eye colour; the AEGAN was unable to generate glasses, and seems to have been confused by them; male-presenting faces are of poorer quality, likely due to their apparent under-representation in the dataset.

\subsection{Interpolation}
GANs famously allow for interpolation between generated samples by interpolating between the latent vectors used to generate those samples. Because AEGAN is bijective, it also allows for the interpolation between \textit{real} samples by encoding those samples as latent vectors, interpolating between the vectors, and generating samples from those interpolated vectors. This process is illustrated in Figure \ref{fig:interpolation}. The AEGAN is able to smoothly interpolate between hair colours, poses, and face and eye shapes. Some issues are plainly visible, however. Eye colour is uniformly green, and several samples were reconstructed with open mouths instead of closed mouths. 

\section{Discussion}
\subsection{Bijective mapping}
Training a GAN to learn a bijective mapping is an elegant solution to the problems of training instability and mode collapse. The Generator in a traditional GAN is never directly exposed to the dataset; indeed, it's remarkable that they work at all. By including a reconstruction component in training, the AEGAN's generator can learn directly from the dataset instead of aimlessly feeling about its way in the dark that is pixel-space, making training much more stable. The reverse mapping, from images to latent space, forces the AEGAN to avoid mode collapse without relying on batch-independence-breaking techniques such as batch normalization or minibatch discrimination.

\subsection{Limitations}
The primary limitation of the AEGAN technique is that it effectively doubles the size of the network, from two sub-networks to four, requiring more time and resources to train. The primary limitation of this particular experiment is that the GAN implementation (https://github.com/ConorLazarou/AEGAN-keras) is fairly simplistic due to time and resource constraints. Hyperparamers, such as number and depth of layers, values of the reconstruction weights $\lambda_{rx}$ and $\lambda_{rz}$, and curriculum methods that would decrease those weights over time were not extensively explored. Further, batch normalization, a staple of GAN implementation, was not used due to an issue in the TensorFlow.Keras implementation, and more sophisticated techniques such as the Wasserstein loss \cite{arjovsky2017wasserstein}, the progressive growing technique \cite{karras2017progressive}, and conditionality \cite{mirza2014conditional} were not applied.

\subsection{Applications and Further Work}
Despite the above shortcomings, there is nothing in the AEGAN formulation that prevents it from being used in conjunction with those techniques, each of which may be an exciting excursion in the GAN landscape. Indeed, I expect combining the AEGAN technique with leading GAN techniques such as StyleGAN would prove very fruitful. Another avenue of exploration is supplementing the multivariate normal distribution used for the latent space with binomial and multinomial elements, as was done in \cite{makhzani2015adversarial}. Such a conditional training scheme may result in an interpretable latent space. One can imagine commercial applications where a potential customer's face is mapped to a vector in latent space, the vector is altered to include sunglasses or some other product, and the new vector is used to produce an image of the customer's face with the product. Obviously this is a silly example, but the ability to map a bijective function between a sample domain and its latent space is a powerful technique, the applications of which are ripe for exploration.

\bibliographystyle{unsrt}  
\bibliography{aegan}

\end{document}